\documentclass[twoside]{article}
\usepackage[accepted]{aistats2018}
\usepackage{times}
\usepackage{graphicx} 
\usepackage{subcaption} 

\usepackage{natbib}

\usepackage{algorithm}
\usepackage{algorithmic}
\usepackage{hyperref}

\usepackage{amssymb}
\usepackage{amsmath}
\usepackage{color}

\begin{document}

%

%

\twocolumn[

\aistatstitle{The Effects of Memory Replay in Reinforcement Learning}

\aistatsauthor{ Ruishan Liu \And James Zou }

\aistatsaddress{ Stanford University \And  Stanford University } ]

\begin{abstract} 
Experience replay is a key technique behind many recent advances in deep reinforcement learning. Allowing the agent to learn from earlier memories can speed up learning and break undesirable temporal correlations. Despite its wide-spread application, very little is understood about the properties of experience replay. How does the amount of memory kept affect learning dynamics? Does it help to prioritize certain experiences? In this paper, we address these questions by formulating a dynamical systems ODE model of Q-learning with experience replay. We derive analytic solutions of the ODE for a simple setting. We show that even in this very simple setting, the amount of memory kept can substantially affect the agent's performance---too much or too little memory both slow down learning. Moreover, we characterize regimes where prioritized replay harms the agent's learning. We show that our analytic solutions have excellent agreement with experiments. Finally, we propose a simple algorithm for adaptively changing the memory buffer size which achieves consistently good empirical performance.  
\end{abstract} 

\section{Introduction}
In reinforcement learning (RL), the agent observes a stream of experiences and uses each experience to update its internal beliefs. For example, an experience could be a tuple of (state, action, reward, new state), and the agent could use each experience to update its value function via TD-learning. In standard RL algorithms, an experience is immediately discarded after it's used for an update. Recent breakthroughs in RL leveraged an important technique called experience replay (ER), in which experiences are stored in a memory buffer of certain size; when the buffer is full, oldest memories are discarded. At each step, a random batch of experiences are sampled from the buffer to update agent's parameters. The intuition is that experience replay breaks the temporal correlations and increases both data usage and computation efficiency \cite{lin1992self}.

Combined with deep learning, experience replay has enabled impressive performances in AlphaGo \cite{silver2016mastering}, Atari games \cite{mnih2015human}, etc. Despite the apparent importance of having a memory buffer and its popularity in deep RL, relatively little is understood about how basic characteristics of the buffer, such as its size, affect the learning dynamics and performance of the agent. In practice, a memory buffer size is determined by heuristics and then is fixed for the agent. 

Prioritized experience replay (pER) is a modification of ER whereby instead of uniformly choosing experiences from the buffer to use in update, the agent is more likely to sample experiences that are ``surprising'' \cite{moore1993prioritized} \cite{schaul2015prioritized}. pER is empirical shown to improve the agent's performance compared to the regular ER, but we also lack a good mathematical model of pER.    


\paragraph{Contributions.} In this paper, we perform a first rigorous study of how the size of the memory buffer affects the agent's learning behavior. We develop an ODE model of experience replay and prioritized replay. In a simple setting, we derive analytic solutions characterizing the agent's learning dynamics. These solutions directly quantify the effects of memory buffer size on the learning rate. Surprisingly, even in this simple case with no value function model mismatch, memory size has a non-monotonic effect on learning rate. Too much or too little memory both can slow down learning. Moreover, prioritized replay could also slow down learning. We confirm these theoretical predictions with experiments. This motivated us to develop a simple adaptive experience replay (aER) algorithm to automatically learn the memory buffer size as the agent is learning its other parameters. We demonstrate that aER consistently improves agent's performance.  

\paragraph{Related works.} The memory replay technique has been widely implemented in RL experiments currently and is shown to have a good performance for different algorithms such as actor-critic RL algorithms \cite{wawrzynski2009real}, deep Q-Network (DQN) algorithms \cite{mnih2013playing, mnih2015human}, and double Q-learning algorithms \cite{van2016deep}.
To further make good use of experience, prioritized methods are proposed for RL algorithms \cite{moore1993prioritized, peng1993efficient}.
The main idea of prioritization is to sample transitions that lead to larger value change in RL more frequently.
The probability of selecting an experience is determined by the relative magnitude of the temporal difference error (TD-error). This has been reported to be effective in many experiments \cite{moore1993prioritized, van2013efficient, schaul2015prioritized}.
Measures other than TD-error are also in literature to weight experience; examples include rewards \cite{tessler2016deep} and the transition property \cite{peng2016terrain}.

The performance of RL with experience replay is similar to the batch RL but in an incremental way \cite{lagoudakis2003least, kalyanakrishnan2007batch, ernst2005tree}.
Another approach to reuse data in RL is called model-learning or Dyna architecture, which builds a model to simulate and generate new data \cite{sutton1990integrated, sutton2008dyna}. This method, however, induces both extra computation cost and modeling error for the data.

\section{A Dynamical System Model of Experience Replay} \label{sec:GeneralModel}

In an RL task, an agent takes actions $a$, observes states $x$ and receives rewards $r$ in sequence during its interaction with the environment. 
The goal is to learn a strategy which leads to best possible reward.
A standard learning framework for the agent is to use the action-value function to learn optimal behavior and perform action selection. The optimal action-value $Q (x, a)$ is defined as the maximum expected return when the agent starts from state $x$ and takes first action $a$. It satisfies
\begin{equation} \label{Eq: Q}
\begin{aligned}
Q (x, a) &= \mathbb{E} \left[ \sum_{i=0}^{K} \gamma^i r(x_i, a_i) \Bigg| x_0 = x, a_0  = a \right] \\ 
& = r(x, a) + \gamma \sum_{y \in X} P(x, a, y) \sup_{a' \in A} Q (y,a'),
\end{aligned}
\end{equation}
where $r(x, a)$ is the reward function, $\gamma$ ($0 \leqslant \gamma < 1$) denotes the discount factor, and $P(x, a, y)$ is
the state transition probability kernel, defined as the probability of moving from state $x$ to state $y$ under action $a$.

In practice, the state space is usually large and the function approximation is adopted to estimate the action-value function $Q(x, a; \theta)$; deep Q-Network (DQN) is an example of this approach. At learning step $t$, the commonly-used TD-learning method updates the weight $\theta$ according to
\begin{equation} \label{Eq: thetaorigin}
\begin{aligned}
\theta(t+1) = & \theta(t) + \alpha(t) \bigg[ r(t') + \gamma \cdot \max_{a' \in A} Q[x(t'+1), a'; \theta(t)]  \\
& - Q\left[x(t'), a(t'); \theta(t) \bigg] \right]  \times \nabla_{\theta} Q[x(t'), a(t'); \theta(t)],
\end{aligned}
\end{equation}
where $\alpha (t)$ is the step size. 
Here the data collected at learning step $t'$ is utilized to do the TD update. 
For standard RL algorithms, only the most recent transition is visited and $t'=t$, while for the ER approach, experience data are reused and $t'<t$.

\begin{algorithm}[] \label{code: RLER}
\caption{Reinforcement Learning with Experience Replay} 
\begin{algorithmic}[1] 
\STATE {\bfseries Input:} memory size N, minibatch size m, step size $\alpha$, discount factor $\gamma$, total steps $T$, initial weights $\theta_0$, update policy $\pi_{\theta}$
\STATE Initialize replay memory $\mbox{BUFFER}$ with capacity N
\STATE Observe initial state $x_0$
\FOR{$t = 1$ to $T$} 
\STATE Take action $a_t \sim \pi_{\theta}(x_t)$ 
\STATE Observe $r_t$ and $x_{t+1}$
\STATE Store transition $(x_t, a_t, r_t, x_{t+1})$ in memory $\mbox{BUFFER}$
\FOR{$j=1$ to $m$}
\STATE Sample a transition $(x_i, a_i, r_i, x_{i+1})$ randomly from $\mbox{BUFFER}$
\STATE Compute TD-error \par
$ \ \ \ \ \delta_i = r_i + \gamma \max_{a} Q(x_{i+1}, a; \theta) - Q(x_{i}, a_i; \theta)$
\STATE Update weights $\theta = \theta + \alpha \delta_i \nabla_{\theta} Q(x_i, a_i; \theta)$
\ENDFOR
\ENDFOR  
\end{algorithmic} 
\end{algorithm}

The effect of the memory buffer can not be extracted from the ER algorithm itself, and the hidden mechanism is hard to perceive only with experiments in a black box.
Thus we derive an ODE model to simulate the learning process. General results are obtained numerically and even analytically, confirmed as good matches with experiments. 
This analytic approach enables us to systematically analyze how the replay memory affects the learning process and what is the principle behind it.

The ODE model corresponds to a continuous interpolation of the discrete learning step $t$ ($t = 0, 1, 2, ...$).
This continuous approximation works well when the step size $\alpha$ is not too large, i.e., there is no dramatical change for the weights within a few steps. This criterion is often met in real experiments. More details of the ODE derivation is in the Appendix.

Under the continuous approximation, the dynamic equation for the weights is
\begin{equation} \label{Eq: theta}
\begin{aligned}
\frac{\mathrm{d}\theta(t)}{\mathrm{d}t} = \alpha(t) & \nabla_{\theta} Q[x(t'), a(t'); \theta(t)] \bigg[ - Q[x(t'), a(t'); \theta(t)] \\
& + r(t') + \gamma \cdot \max_{a' \in A} Q[x(t'+1), a'; \theta(t)] \bigg] ,
\end{aligned}
\end{equation}

The agent's state evolution can also be estimated as
\begin{equation} \label{Eq: xt}
\frac{\mathrm{d}x(t)}{\mathrm{d}t} = \int_X P \left[ x(t), a(t), y \right] y \mathrm{d} y
\end{equation}

Here the action $a(t)$ is selected according to a policy $\pi$
\begin{equation} \label{Eq: at}
a(t) = \pi [x(t)]
\end{equation}

Together with Eq. (\ref{Eq: xt}), the state evolving trajectory can be depicted. For instance, with an $\epsilon-$greedy policy, the state movement obeys
\begin{equation}
\begin{aligned}
\frac{\mathrm{d}x(t)}{\mathrm{d}t} = & \int_X y \mathrm{d}y \bigg[ \frac{\epsilon \int_A P \left[ x(t), a, y \right] \mathrm{d} a}{ \int_A \mathrm{d}a} \\
& + (1-\epsilon) P [ x(t), \mathrm{arg}\max_{a \in A} Q[x(t), a; \theta(t)]), y]  \bigg]
\end{aligned}
\end{equation}


In ER, recent transitions are stored in a replay memory with the capacity $N$, and a minibatch of data is randomly chosen for TD-learning. 
The parameter dynamics under experience replay is 
\begin{equation} \label{Eq: theta_memory}
\begin{aligned}
\frac{\mathrm{d}\theta(t)}{\mathrm{d}t} & = \frac{m\alpha(t)}{n(t)} \int_{t-n(t)}^{t} \mathrm{d} t' \nabla_{\theta} Q[x(t'), a(t'); \theta(t)] \bigg[ r(t') \\ 
& + \gamma \cdot \max_{a \in A} Q[x(t'+1), a; \theta(t)] - Q[x(t'), a(t'); \theta(t)] \bigg] 
\end{aligned}
\end{equation}
where $m$ is the minibatch size and $n(t) \leqslant N$ is the memory size. Eq. (\ref{Eq: theta_memory}) can be viewed as giving the expected gradient value of the parameter updates at each time step. 

Now we are able to analyze the learning process, more specifically, the weights $\theta (t)$ and state $x (t)$ as a function of learning step, based on Eq. (\ref{Eq: xt}) and Eq. (\ref{Eq: theta_memory}). 
No experiment is needed and the influence of the memory buffer or other parameters can be analyzed explicitly from the analytical solutions.
Our theoretical model is further validated by experiments based on ER algorithms.

Prioritized replay (pER) proposes to speed up the learning process by sampling the experience transitions according to a non-uniform probability distribution. One commonly used model is to parametrize the probability for selecting transition $i$ as
\begin{equation}
P(i) = \frac{|\delta_i|^{\beta}}{ \sum_j |\delta_j|^{\beta}},
\end{equation}
where $\beta$ is a constant exponent and $\delta_i = r_i + \gamma \max_{a} Q(s_{i+1}, a; \theta) - Q(s_{i}, a_i; \theta)$ is the TD-error. The only difference between ER and pER is in how to sample experiences. 

Taking $\beta=2$ as an instance, the dynamic equation for weights under pER is given by
\begin{equation} \label{Eq: theta_memory_pri}
\begin{aligned}
\frac{\mathrm{d}\theta(t)}{\mathrm{d}t} = & \frac{m\alpha(t) \int_{t-n}^{t} \delta^3 (t') \nabla_{\theta} Q[x(t'), a(t'); \theta(t)] \mathrm{d} t' }{\int_{t-n}^{t} \delta^2(t') \mathrm{d} t'}
\end{aligned}
\end{equation}
where $\delta(t') = r(t') + \gamma \cdot \max_{a \in A} Q[x(t'+1), a; \theta(t)] - Q[x(t'), a(t'); \theta(t)]$.

\section{Analysis of Memory Effects in a Simple Setting} \label{sec:ToyExample}
Starting from a toy game we call LineSearch, we analytically solve the ODEs (\ref{Eq: theta_memory}) and (\ref{Eq: theta_memory_pri}) to get the learning dynamics and quantify the effects of memory.
We further characterize settings when pER helps or hinders learning tasks compared to ER.
Finally, we show that our theoretical predictions have excellent agreement with experiments. 

\paragraph{Model setup.} 
We first define a simple game LineSearch, for which the space of agent state $x$ is one dimensional, the reward function is linear $r(x) = \beta_1 x + \beta_2$, and the action is binary $a \in \{v, -v\}$, where $v$ is a constant.
In a transition, the next state is determined by adding the action value to the current state, i.e., $x(t+1) = x(t) + a(t)$.
When the discount factor $\gamma$ is set as 0 (non-zero $\gamma$ setting gives similar agent's behavior and will be addressed later), the real action-value function is 
\begin{equation} \label{Eq: Qreal}
\begin{aligned}
Q_{\mathrm{real}}(x,a) & =  r(x+a) \\
& = \beta_1 \cdot (x+a) + \beta_2
\end{aligned}
\end{equation} 
When there is no model mismatch, the action-value function of the agent is
\begin{equation} \label{Eq: Qagent}
Q_{\mathrm{agent}}(x, a; \theta) = \theta_{1} \cdot (x+a) + \theta_{2},
\end{equation} 
At $t=0$, $\theta_1$ and $\theta_2$ are randomly initialized. As the agent performs TD update, we are interested in how quickly the  $\theta$'s approach the true $\beta$'s. A natural evaluation metric is $\Delta \theta_1$ and $\Delta \theta_2$ defined as
\begin{equation}
\Delta \theta_1 = \theta_1 - \beta_1 \mbox{  and  } \Delta \theta_2 = \theta_2 - \beta_2
\end{equation}
The agent learns well if $\Delta \theta_1$ and $\Delta \theta_2$ approach $0$, and performs badly when $\Delta \theta_1$ or $\Delta \theta_2$ is large.
Under a greedy policy, the evolution of the agent's state is
\begin{equation} \label{Eq: x}
\frac{\mathrm{d}x(t)}{\mathrm{d}t} = \frac{\theta_1}{|\theta_1|} v,
\end{equation}

\begin{figure*}[h]
\centering
\begin{subfigure}[t]{.32\textwidth}
\centering
  \includegraphics[width=1\linewidth]{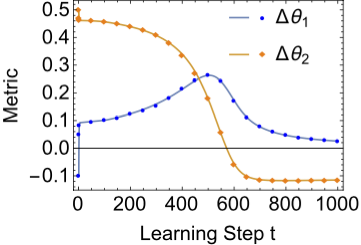}
  \caption{ER.}
  \label{fig: 2DLearning}
\end{subfigure} \hfill
\begin{subfigure}[t]{.32\textwidth}
\centering
  \includegraphics[width=1\linewidth]{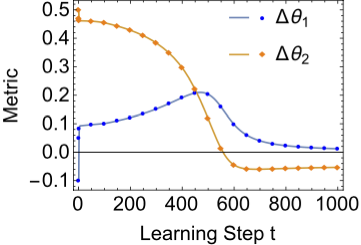}
  \caption{pER.}
  \label{fig: 2DPriLearning}
  \end{subfigure}\hfill
\begin{subfigure}[t]{.315\textwidth}
\centering
  \includegraphics[width=1\linewidth]{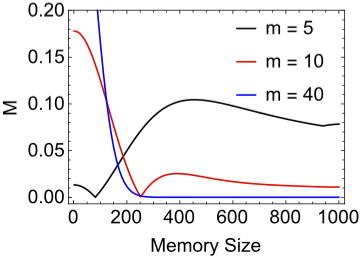}
  \caption{$M$ vs. memory size for ER.}
  \label{fig: 2DMetricVsSize}
\end{subfigure} \hfill
\begin{subfigure}[t]{.31\textwidth}
\centering
  \includegraphics[width=1\linewidth]{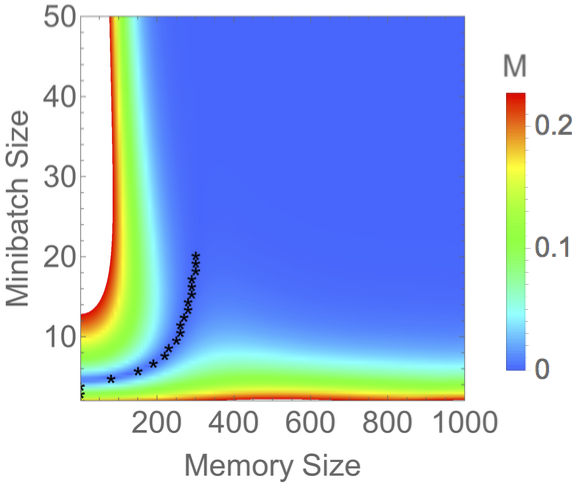}
  \caption{ER with $\gamma = 0$.}
  \label{fig: 2D2D}
\end{subfigure} \hfill
\begin{subfigure}[t]{.32\textwidth}
\centering
  \includegraphics[width=1.03\linewidth]{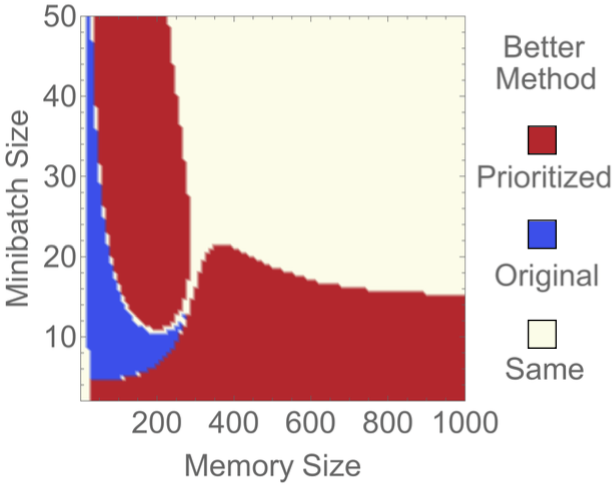}
  \caption{Compare ER and pER}
  \label{fig: 2DCompSign}
  \centering
\end{subfigure} \hfill
\begin{subfigure}[t]{.31\textwidth}
\centering
  \includegraphics[width=1\linewidth]{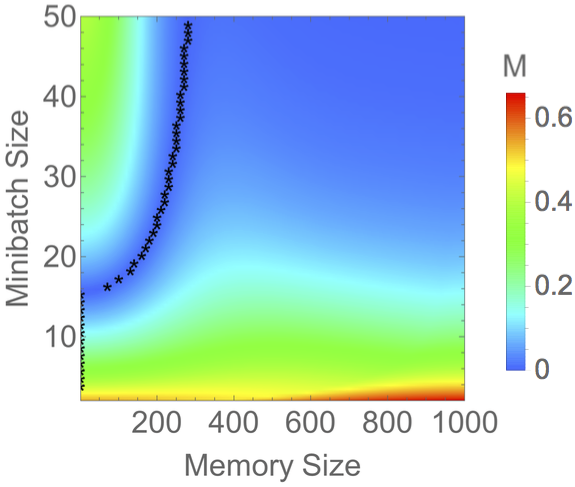}
  \caption{ER with $\gamma = 0.7$.}
  \label{fig: 2DGamma07}
\end{subfigure}
\caption{(a,b) Learning curves for two metrics $\Delta \theta_1 = \theta_1 - \beta_1$ and $\Delta \theta_2 = \theta_2 - \beta_2$, where $\theta_{1(2)}$ and $\beta_{1(2)}$ are the weights of the agent and the real weights, respectively. The scattered blue dots and orange squares represent the experimental results for $\Delta \theta_1$ and $\Delta \theta_2$, using (a) ER algorithm and (b) pER algorithm. The blue and orange curves are the numerical solutions for $\Delta \theta_1$ and $\Delta \theta_2$ based on our theoretical model. \emph{For both ER and pER, our theoretical predictions have excellent agreement with the experimental results on the learning dynamics.}  
(c) Dependence of final absolute metric sum $M=|\Delta \theta_1 (t=1000)|+|\Delta \theta_2 (t=1000)|$ on memory size for different minibatch sizes. Note that smaller $M$ stands for better performance.  Here we use the original setting when the discount factor $\gamma=0$; $m$ indicate batch size. 
(d,f) Contour plot of measure $M$ as a function of memory size and minibatch size, with the discount factor (d) $\gamma = 0$ and (f) $\gamma=0.5$. The stars denote the optimal memory sizes given minibatch values. The plots in (c) corresponds to the situations in (d) when the minibatch size is 5, 10, and 40.
(e) The red (blue) region stands for the situations when pER (ER) works better. The rest white region is the situations when the two settings behave similarly. More precisely, the absolute difference of $M$ for the ER and pER is less than $1.5 \times 10^{-3}$ in the white area.  
Here (c-f) are plotted based on theory predictions and also fit experiments well, similar to (a,b). }
\label{fig: 2DLearnings}
\end{figure*}

With the initial state denoted as $x_0$, the evolution of the two metrics is derived from Eqs. (\ref{Eq: theta_memory})-(\ref{Eq: x}) as
\begin{equation} \label{Eq: dtheta1}
\frac{\mathrm{d}\Delta \theta_1(t)}{\mathrm{d}t} = - (b_{10} + b_{11} t + b_{12} t^2) \Delta \theta_1(t) - (b_{20} + b_{21} t) \Delta \theta_2(t)
\end{equation}
\begin{equation} \label{Eq: dtheta2}
\frac{\mathrm{d}\Delta \theta_2(t)}{\mathrm{d}t} = - (b_{20} + b_{21} t) \Delta \theta_1(t) - b_{22} \Delta \theta_2(t),
\end{equation}
where $b_{10} = m\alpha(n^2 v^2/3 + x_0^2 - n v x_0)$, $b_{11} = m\alpha(2 v x_0-n v^2)$, $b_{11} = m\alpha v^2$, $b_{20} = m\alpha (x_0 - nv/2)$, $b_{21} = m\alpha v$, and $b_{22} = m\alpha$. Here our discussion is set in the  $\theta_1 > 0$ region, similar study could be carried out when $\theta_1 < 0$.
Throughout this section, we choose the initial state $x_0 = -5$, the action amplitude $v = 0.01$, and the initial metrics $\Delta \theta_1^0 = -0.1$ and  $\Delta \theta_2^0 = 0.5$.

The learning process in theory can then be calculated based on the dynamic equations of the metrics, i.e., Eq. (\ref{Eq: dtheta1}) and Eq. (\ref{Eq: dtheta2}). It should be noted that the weights $\theta$ are obtained at the same time as $\theta_1 = \Delta \theta_1 + \beta_1$ and  $\theta_2 = \Delta \theta_2 + \beta_2$.  
The detailed analytical solution and discussion for the ODEs Eq. (\ref{Eq: dtheta1}) and Eq. (\ref{Eq: dtheta2}) are given in Appendix.

The theoretical learning curves for both RL and pRL settings are depicted in Fig. \ref{fig: 2DLearning} and Fig. \ref{fig: 2DPriLearning}, with the minibatch size $m$ being $5$ and the step size $\alpha$ being $0.01$. The two metrics $\Delta \theta_1$ and $\Delta \theta_2$ are represented by the solid blue and orange curve, respectively.

To demonstrate the validity of our theoretical solution, we also performed experiments on LineSearch following ER and pER algorithms. For illustration, we plot the experimental results in Fig. \ref{fig: 2DLearning} and Fig. \ref{fig: 2DPriLearning}, where the blue dots stand for $\Delta \theta_1$ and the orange square denotes $\Delta \theta_2$. Our theoretical prediction has excellent agreement with experiment results.

\paragraph{Effects of memory size.} 
By solving the ODEs, we found that the memory setting has a non-monotonic effect on the RL performance.
We are able to extract the mechanism behind this phenomenon from the analytic expressions.

With $\theta_2$ being fixed as the real value $\theta_2 \equiv \beta_2$, the metric $\Delta \theta_1(t)$ is solved analytically
\begin{equation} \label{Eq: 1DthetaMain}
\begin{aligned}
\Delta \theta_1(t) = & \Delta \theta_1^0 e^{ - m\alpha \left[ \frac{v^2}{3} t^3 + \frac{v(2x_0-N v)}{2} t^2 \right]} \\ 
& \cdot e^{ - m\alpha \left[ \left( x_0^2-N v x_0 + \frac{N^2 v^2}{3} \right) t  -\frac{1}{18} N^2 v \left(N v-9 x_0\right) \right] },
\end{aligned}
\end{equation}
The metric converges exponentially to 0, but the rate of convergence (the exponent) has non-monotonic dependence on the memory size $N$. 

In contrast, when $\theta_1$ is fixed to the correct value, $\theta_1 \equiv \beta_1$, the metric $\Delta \theta_2(t)$ evolves according to
\begin{equation} \label{Eq: 1Dtheta2Main}
\Delta \theta_2(t) = \Delta \theta_2^0 e^{-m\alpha t}
\end{equation}
In this case, the choice of memory size has no effect on the learning behavior.  
Detailed explanation and analysis are in Appendix.

Now we turn to a more general situation when the updates of the two weights are coupled together. 
To examine the performance, we define a measure $M=|\Delta \theta_1 (t=1000)|+|\Delta \theta_2 (t=1000)|$, i.e., the sum of the two metrics absolute values at the learning step 1000, the end of the game. The smaller the measure $M$ is, the better the agent learns. 
Fig. \ref{fig: 2D2D} plots the dependence of $M$ on the memory size $N$ and the minibatch size $m$, with the step size $\alpha=10^{-3}$. 


The learning performance is affected non-monotonically by the memory size for $m < 20$, while a monotonic relation is observed for $m>20$, as shown in Fig. \ref{fig: 2D2D}.
For example, an optimal memory size around $250$ exists for $m=10$, denoted by the red curve in Fig. \ref{fig: 2DMetricVsSize}. In contrast, the measure $M$ experiences a monotonic decrease with the growth of memory size for $m=40$, plotted by the blue curve in Fig. \ref{fig: 2DMetricVsSize}.

The influence of the memory setting in RL arises from the trade-off between the overshooting and the weight update.
Here the term overshooting describes the phenomenon when some of the weights are updated in the wrong direction. 
For example, in Fig.~\ref{fig: 2DLearning}, $\theta_1$ actually moves further away from $\beta_1$ during times 200 to 500; $\theta_2$ also overshoots at $T = 600$ and incurs negative bias. 
We first address the settings with small minibatches.
When the memory size is also small, the learning process is more likely to overshoot because of the limited memory capacity.
When the replay memory is enlarged, the overshooting effect is mitigated. 
With the increase of the memory size, the averaged weight update first becomes slowly then slightly accelerates.
The balance between these two contributions leads to the non-monotonic nature.
When $m$ is large,
there is still a trade-off between overshooting and increasing weight update. However, the latter can not counteract the former because of the quick convergence induced by the large TD update. Analytical expressions and numerical results are combined for illustrations in Appendix.

\paragraph{Performance of prioritized replay.} 
We further compare pER and ER, and discuss how the memory buffer affects pER based on our theoretical model.
Fig. \ref{fig: 2DPriLearning} plots the learning curve for pER, which exhibits a similar property as for ER in Fig. \ref{fig: 2DLearning} .
We compare the performance of RL and pRL algorithms in Fig. \ref{fig: 2DCompSign}, where blue (red) regions stand for cases when ER (pER) is better, and white areas represent situations when the two algorithms perform similarly.
It is shown that pER performs relatively worse when the memory size is small, particularly when the minibatch size is not large. 
This is also attributed to the trade-off between the overshooting and quick weight update. 
For small memory size, the overshooting effect is more serious under the prioritized sampling, while for a large memory, the prioritized agents update the weight quicker which leads to a faster convergence. Demonstrations are given in Appendix.
\paragraph{Nonzero discount factor.}
The discount factor is set as 0 in the previous subsections for simplicity. Here we show that the learning dynamics with $\gamma>0$ is qualitatively similar to the case when $\gamma=0$ in the LineSearch game.
With a nonzero discount factor $\gamma$, i.e., considering the long-term effect, the real action-value function under the greedy policy is 
\begin{equation} \label{Eq: QrealNonzero}
\begin{aligned}
Q_{\mathrm{real}}(x,a) & = \mathbb{E} \bigg[ \sum_{i=0}^{K} \gamma^i [\beta_1 (x_0 + a_0 + i |\beta_1| v) \\ 
& \ \ \ \ \ \ \ \ \ \ \ \ \ \ \ \ \ \ \ \ \ \ \ \ \ + \beta_2] \Bigg| x_0 = x, a_0 = a \bigg] \\
& = \frac{1-\gamma^{K+1}}{1-\gamma} [\beta_1 (x+a) + \beta_2] \\
& \ \ \ \ \ +  v |\beta_1| \left[ \frac{\gamma-\gamma^{K+1}}{(1-\gamma)^2} - \frac{K\gamma^{K+1}}{1-\gamma} \right] \\
& \approx \frac{\beta_1}{1-\gamma} (x+a) + \frac{\beta_2}{1-\gamma} + \frac{\gamma v |\beta_1|}{(1-\gamma)^2},
\end{aligned}
\end{equation}
where $K$ stands for the total training steps afterwards before the game ends. The approximation in Eq. (\ref{Eq: QrealNonzero}) is valid for large $K$ and $\gamma < 1$. For instance, with the discount factor $\gamma = 0.9$, the contribution of the term $\gamma^K$ after 100 steps is $\gamma^{100} = 0.00003$.

When there is no model mismatch, the action-value function of the agent is
\begin{equation} \label{Eq: QagentNonzero}
Q_{\mathrm{agent}}(x, a; \theta) = \theta_{1} \cdot (x+a) + \theta_{2},
\end{equation} 

Then the evolution of the two weights are derived together with Eqs. (\ref{Eq: theta_memory})-(\ref{Eq: x}) as
\begin{equation} \label{Eq: dtheta1Nonzero}
\begin{aligned}
\frac{\mathrm{d} \theta_1(t)}{\mathrm{d}t} = & - (b_{10} + b_{11} t + b_{12} t^2) \left[ (1-\gamma) \theta_1(t) - \beta_1 \right] \\
& - (b_{20} + b_{21} t)  [ (1-\gamma) \theta_2(t) - \gamma v |\theta_1 (t)| - \beta_2 ]
\end{aligned}
\end{equation}
\begin{equation} \label{Eq: dtheta2Nonzero}
\begin{aligned}
\frac{\mathrm{d} \theta_2(t)}{\mathrm{d}t} = & - (b_{20} + b_{21} t) \left[ (1-\gamma) \theta_1(t) - \beta_1 \right] \\
& - b_{22} \left[ (1-\gamma) \theta_2(t) - \gamma v |\theta_1 (t)| - \beta_2 \right],
\end{aligned}
\end{equation}
where $b_{10} = m\alpha(n^2 v^2/3 + x_0^2 - n v x_0)$, $b_{11} = m\alpha(2 v x_0-n v^2)$, $b_{11} = m\alpha v^2$, $b_{20} = m\alpha (x_0 - nv/2)$, $b_{21} = m\alpha v$, and $b_{22} = m\alpha$. The learning curve is attached in Appendix.

We find that the ODEs for $\gamma > 0$ have similar form as $\gamma = 0$. Correspondingly, the results obey similar principles, as shown in Fig. \ref{fig: 2DGamma07}.
Here the step size is $\alpha=10^{-3}$, the real weights $\theta_1^r$ and $\theta_2^r$ are 0.1 and 0.5, and the initial weights $\theta_1^0$ and $\theta_2^0$ are 0 and 1.


\begin{algorithm}[t] \label{code: RLER}
\caption{Adaptive Memory Size Reinforcement Learning with Experience Replay} 
\begin{algorithmic}[1] 
\STATE {\bfseries Input:} initial memory size $N_0$, minibatch size m, step size $\alpha$, discount factor $\gamma$, total steps $T$, initial weights $\theta_0$, update policy $\pi_{\theta}$, number of checked oldest transitions $n_{\mathrm{old}}$ and memory adjustment internal $k$.
\STATE Initialize replay memory $\mbox{BUFFER}$ with capacity $N=N_0$ and set $|\delta_{\mathrm{old}}| = 0$
\STATE Observe initial state $x_0$
\FOR{$t = 1$ to $T$} 
\STATE Take actions, observe, store transitions and do TD updates as in ER algorithm
\IF {mod(t, k) = 0 and memory $\mbox{BUFFER}$ is full}
\STATE Compute $|\delta_{\mathrm{old}}|' = \sum_{i = t-N+1}^{t-N+n_{\mathrm{old}}} | r_i + \gamma \max_{a} Q(x_{i+1}, a; \theta) - Q(x_{i}, a_i; \theta)|$ 
\IF {$|\delta_{\mathrm{old}}|' > |\delta_{\mathrm{old}}|$ or $N = k$}
\STATE Enlarge the memory $N = N + k$
\STATE $|\delta_{\mathrm{old}}| = |\delta_{\mathrm{old}}|' $
\ELSE
\STATE Shrink the memory $N = N - k$, delete the oldest $k$ transitions in $\mbox{BUFFER}$
\STATE Compute $|\delta_{\mathrm{old}}| = \sum_{i = t-N+k+1}^{t-N+k+n_{\mathrm{old}}} | r_i + \gamma \max_{a} Q(x_{i+1}, a; \theta) - Q(x_{i}, a_i; \theta)|$
\ENDIF
\ENDIF
\ENDFOR
\end{algorithmic} 
\end{algorithm}

\section{Adaptive Memory Size Algorithm}
The analysis from the previous section motivated us to develop a new algorithm that allows the agent to adaptively adjust the memory size while it is learning other parameters. 

\paragraph{Intuition.} The lesson from Section~\ref{sec:ToyExample} is that an useful adaptive algorithm should increase the memory capacity when the overshooting effect dominates and shrink the memory buffer if the weight update becomes too slow.

We use the \emph{change} in absolute TD error of the oldest memories in the buffer as a proxy for whether the agent is overfitting to more recent memories. 
The intuition is that if the TD error magnitude for the oldest transitions in the buffer starts to increase---i.e. the old data violate the Bellman equation more severely as the agent learns---then this is a sign that the agent might be overshooting and overfitting for the more recent experiences. In this case, we increase the memory buffer to ensure that the older experiences are kept longer to be used for future updates. On the other hand, if the TD error magnitude for the oldest transitions in the buffer starts to decrease over time, then the older memories are likely to be less useful and the agent decreases the memory buffer to accelerate the learning process.

\paragraph{Algorithm description.} The memory size is adaptively changed according to the TD error magnitude change of the oldest transitions, characterized by $|\delta_{\mathrm{old}}|' - |\delta_{\mathrm{old}}|$. Here $|\delta_{\mathrm{old}}|$ and $|\delta_{\mathrm{old}}|'$ are defined as the sum of the absolute TD errors of the old $n_{\mathrm{old}}$ transitions in memory, where $n_{\mathrm{old}}$ is a hyperparameter which denotes the number of old data we choose to examine.
$|\delta_{\mathrm{old}}|$ is first calculated. After $k$ steps, we derive $|\delta_{\mathrm{old}}|'$ and compare it with $|\delta_{\mathrm{old}}|$. More specifically, every $k$ steps, if the change of absolute TD error magnitude sum for the old transitions decreases, i.e., $|\delta_{\mathrm{old}}|' < |\delta_{\mathrm{old}}|$, the memory shrinks, otherwise increases, as given in Algorithm 2, denoted as aER.

\paragraph{Performance on LineSearch.} 

\begin{figure}[]
\centering
\begin{subfigure}[t]{.3\textwidth}
\centering
  \includegraphics[width=1\linewidth]{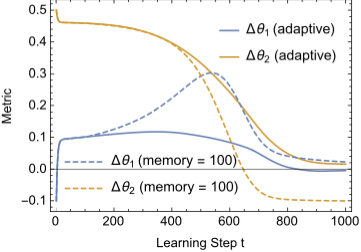}
  \caption{Learning curve for the LineSearch game, when the memory is learned adaptively from 100 (solid) vs. fixed at 100 (dashed). Metric equal to 0 is optimal. }
  \label{fig: AdaptiveLinear}
\end{subfigure} \hfill
\begin{subfigure}[t]{.31\textwidth}
\centering
  \includegraphics[width=1\linewidth]{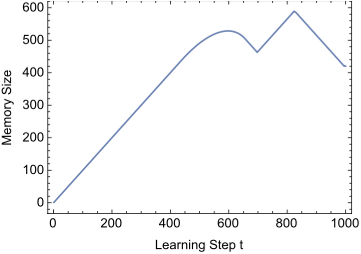}
  \caption{Adaptive memory for the LineSearch game starting from memory size of 100. }
  \label{fig: AdaptiveLinearMemory}
\end{subfigure} \hfill
\caption{Learning curve and adaptive memory for LineSearch and CartPole.}
\label{fig: adaptive}
\end{figure}

\begin{figure*}[h]
\centering
\begin{subfigure}[t]{.3\textwidth}
\centering
  \includegraphics[width=1\linewidth]{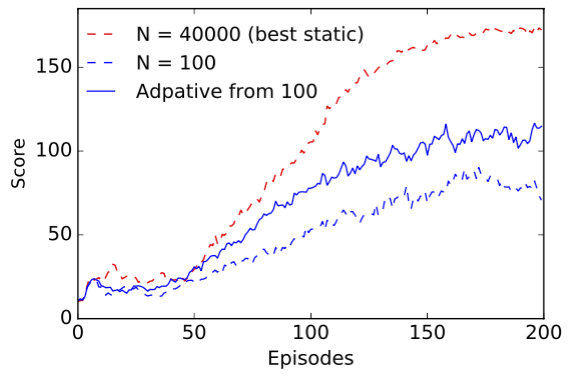}
  \caption{Learning curve for CartPole, when the memory is adaptively learned starting from 100, and fixed as 100 and 40000, i.e., no experience is discarded.}
  \label{fig: Cartpole}
\end{subfigure} \hfill
\begin{subfigure}[t]{.31\textwidth}
  \includegraphics[width=1\linewidth]{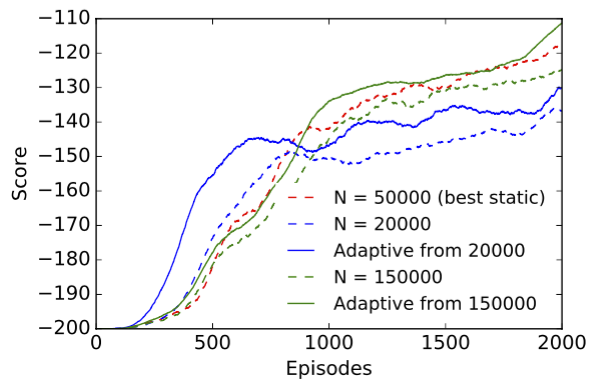}
  \caption{Learning curve for MountainCar, when the memory is fixed as 20000, 50000 and 150000, and adaptively learned starting from 20000 and 150000.}
  \label{fig: MountainCar}
\end{subfigure} \hfill
\begin{subfigure}[t]{.31\textwidth}
  \includegraphics[width=1\linewidth]{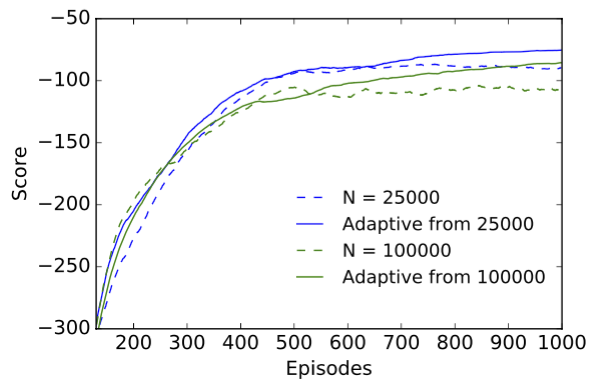}
  \caption{Learning curve for Acrobot, when the memory is fixed as 25000 and 100000, and adaptively learned starting from 25000 and 100000.}
  \label{fig: Acrobot}
\end{subfigure} \hfill
\begin{subfigure}[t]{.3\textwidth}
\centering
  \includegraphics[width=1\linewidth]{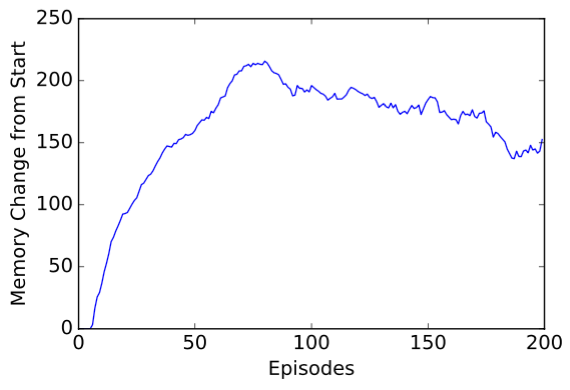}
  \caption{Adaptive memory change for CartPole starting from memory size of 100.}
  \label{fig: CartpoleMemory}
\end{subfigure} \hfill
\begin{subfigure}[t]{.32\textwidth}
\centering
  \includegraphics[width=1\linewidth]{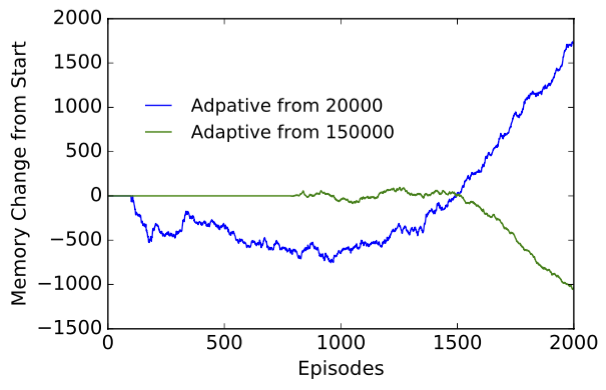}
  \caption{Adaptive memory compared to the initial size 20000 and 150000 for MountainCar. }
  \label{fig: MountainCarMemory}
  \centering
\end{subfigure} \hfill
\begin{subfigure}[t]{.31\textwidth}
\centering
  \includegraphics[width=1\linewidth]{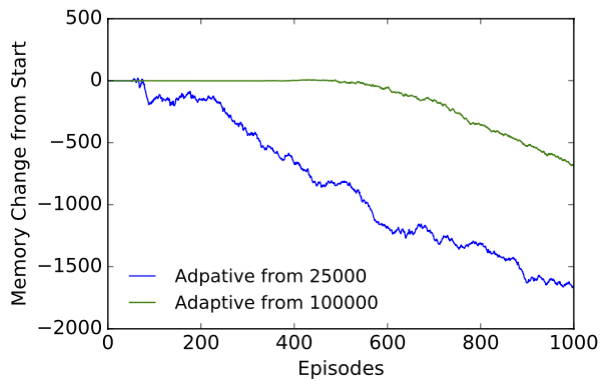}
  \caption{Adaptive memory compared to the initial size 25000 and 100000 for Acrobot.}
  \label{fig: AcrobotMemory}
\end{subfigure}
\caption{(a-c) Learning curves for CartPole, MountainCar and Acrobot, which are averaged for 100 new games. (d-f) Adaptive memory change from the initial size, corresponding to (a-c).}
\end{figure*}

We first analyze how aER works for the LineSearch game. 
With the minibatch size $m=10$ and the step size $\alpha = 10^{-3}$, the agent adaptively adjusts its memory capacity from 100 as depicted in Fig. \ref{fig: AdaptiveLinearMemory}.
Compared to the setting with fixed memory size as $100$, the agent updates weights more effectively and the overshooting effect is mitigated, indicated by Fig. \ref{fig: AdaptiveLinear}.

\paragraph{Performance on OpenAI Games with DQN.} 
We further evaluated the algorithm on three standard RL benchmarks that we downloaded from OpenAI Gym, which are CartPole, MountainCar and Acrobot.

We used DQN with fully connected neural network (NN) for the value function approximation, where the NN is one layer for CartPole and two layers for MountainCar and Acrobot.
Here we randomly initialize the weights, and set the minibatch size $m = 50$ and memory adjustment internal $k = 20$.
The checked old transitions cover half of the initial memory size, out of which we randomly sample 50 (1000) experiences to approximate $|\delta_{\mathrm{old}}|'$ for CartPole (MountainCar and Acrobot).
The discount factor $\gamma$ is set to be $0.9$ for CartPole and $0.99$ for MountainCar and Acrobot. The step size $\alpha$ is $2 \times 10^{-5}$, $6 \times 10^{-4}$ and $10^{-3}$ for CartPole, MountainCar and Acrobot.
 
The adaptive memory algorithm achieved better performance compared to having a fixed-size replay buffer in all three games.


First, for the CartPole game which starts with an initial memory size of 100, the agent speeds up its learning from adaptive adjustments of the memory size, as shown in Fig. \ref{fig: Cartpole} and Fig. \ref{fig: CartpoleMemory} where each curve is averaged for 100 new games. Here a larger static memory is always better and the best performance is achieved when no experience is discarded, corresponding to the full size $40,000$. With the aER, the agent correctly learns to increase its memory size from a small initial size of 100. We note that in 71\% of the trials the aER outperforms the averaged static memory strategy.  

In the second example, MountainCar, we found that the optimal fixed memory size is around 50,000, as indicated by Fig. \ref{fig: MountainCar} and Fig. \ref{fig: MountainCarMemory} where the result is averaged for 100 new games. Both smaller and larger static memory sizes reduce learning. In this setting, aER learns to increase the memory size if the initial memory is small and it also learns to decrease the capacity if the initial memory is too large. 
62\% and 86\% trials with aER outperforms the averaged fixed memory results for the initial size of 20,000 and 150,000, respectively.
We note that the dynamical change of memory size from 150,000 also enables the agent to even perform better than the best static result.

Last, aER also demonstrates good performance in the Acrobot game where a relatively smaller fixed memory is preferred, as plotted in Fig. \ref{fig: Acrobot} and Fig. \ref{fig: AcrobotMemory} where each curve is averaged for 100 new games. When the initial buffer size is 25,000 and 100,000, aER outperforms the averaged fixed memory approach in 100\% and 95\% of the experiments, respectively. Here the agent learns to accelerate its learning by dynamically decreasing the memory size.

\paragraph{Additional considerations} 
Little extra computation cost is needed to carry out the aER. Taking the CartPole game for example, only every 20 steps, we need to compute one or two forward passes of neural network without backpropogation. The examined batch is in the same size as the sampling minibatch for TD update.

The simple aER algorithm has a tendency to shrink the memory, but already shows good performance in experiments. The goal of the TD learning process is to diminish the TD error amplitude for all data, so the updated $|\delta_{\mathrm{old}}|'$ is more likely to be less than $|\delta_{\mathrm{old}}|$ in average sense. 
One possible solution is to change the criterion for shrinking the memory buffer to be $|\delta_{\mathrm{old}}|' < |\delta_{\mathrm{old}}| - \epsilon$, where $\epsilon$ could be a predefined constant, or learned online such as from the averaged TD error amplitude change through the whole dataset and the previous $|\delta_{\mathrm{old}}| - |\delta_{\mathrm{old}}|'$ value.

\vspace{-0.1cm}
\section{Discussion} 
\vspace{-0.1cm}

Our analytic solutions, confirmed by experiments, demonstrate that the size of the memory buffer can substantially affect the agent's learning dynamics even in very simple settings. Perhaps surprisingly, the memory size effect is non-monotonic even when there is no model mismatch between the true value function and the agent's value function. Too little or too much memory can both slow down the speed of agent's learning of the correct value function. We developed a simple adaptive memory algorithm which evaluates the usefulness of the older memories and learns to automatically adjust the buffer size. It shows consistent improvements over the current static memory size algorithms in all four settings that we have evaluated. There are many interesting directions to extend this adaptive approach. For example, one could try to adaptively learn a prioritization scheme which improves upon the prioritized replay. This paper focused on simple settings in order to derive clean, conceptual insights. Systematic evaluation of the effects of memory buffer on large scale RL projects would also be of great interest.

\newpage
\appendix

\section{Analytic solutions of the ODEs of the learning dynamics} \label{sec:ODE_solutions}
Utilizing our model, we are able to analyze the learning properties and the influence of the replay memory systematically. 
How to choose replay memory settings, such as memory size and minibatch size, are further discussed. 

\subsection{Solutions for 1D weights}\label{subsec:1D}
In this subsection, we start with the simplest case when the weight are one-dimensional to get some basic intuition and to prepare for more complexed cases which will be addressed later.

\paragraph{Fix the intercept $\theta_2$} 
We first consider the setting when the intercept $\theta_2$ in Eq. (\ref{Eq: Qagent}) is fixed. 
In this case, the approximated action-value function of the agent $Q_{\mathrm{agent}}$ is
\begin{equation} \label{Eq: 1Dtheta}
Q_{\mathrm{agent}}(x, a; \theta) = \theta \cdot (x+a) + \beta_2
\end{equation}
Note that the real action-value function is $Q_{\mathrm{real}}(x, a; \theta) = \beta_1 \cdot (x+a) + \beta_2 $. 
The weight is initialized to be $\theta^0$, and the initial metric is $\Delta \theta^0 = \theta^0 - \beta_1$ correspondingly.

The metric as a function of learning step $\Delta \theta (t)$ is calculated analytically as
\begin{equation} \label{Eq: 1Dtheta}
\Delta \theta(t) = \Delta \theta^0 e^{-k(t)},
\end{equation}
where the exponent $k(t)$ is given by
\begin{subequations} \label{Eq: 1Dk}
\begin{equation} \label{Eq: 1Dk1}
k(t) = m\alpha[\frac{v^2}{9} t^3 + \frac{x_0 v}{2} t^2 + x_0^2 t]  \ \ \ \ \ (t \leqslant N)
\end{equation}
\begin{equation} \label{Eq: 1Dk2}
\begin{aligned}
k(t) = m\alpha \bigg[ & \frac{v^2}{3} t^3 + \frac{v(2x_0-N v)}{2} t^2 -\frac{1}{18} N^2 v \left(N v-9 x_0\right)\\  
&  + \left( x_0^2-N v x_0 + \frac{N^2 v^2}{3} \right) t  \bigg] \ \ \ \ \ \ \ \ (t > N)
\end{aligned}
\end{equation}
\end{subequations}
The learning process consists of two parts. 
In the beginning, The exponent $k$ gradually grows from 0, described by Eq. (\ref{Eq: 1Dk1}). 
After the replay memory gets full, the memory buffer becomes a sliding window and the exponent evolves according to Eq. (\ref{Eq: 1Dk2}).
The metric $\Delta \theta$ approaches the desired value 0 exponentially and the exponent is a cubic function of the learning step $t$.

\begin{figure}[h]
\centering
\begin{subfigure}{.4\textwidth}
  \centering
  \includegraphics[width=0.9\linewidth]{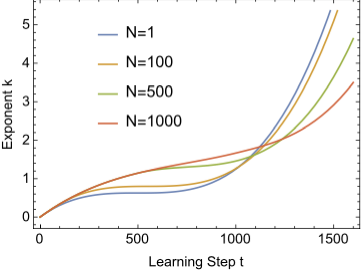}
  \caption{Dependence of exponent $k$ on learning step $t$.}
  \label{fig: 1DLearninga}
\end{subfigure} \hfill
\begin{subfigure}{.4\textwidth}
  \centering
  \includegraphics[width=0.93\linewidth]{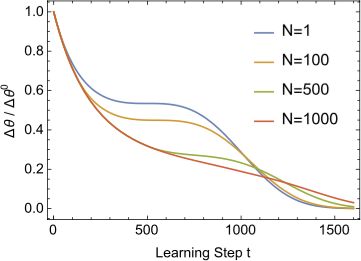}
  \caption{Metric $\Delta \theta / \Delta \theta^0$ vs. learning step $t$.}
  \label{fig: 1DLearningb}
\end{subfigure}
\caption{Learning curve for the exponents $k$ and the metric $\Delta \theta / \Delta \theta^0 = e^{-k}$. The memory sizes $N$ are 1, 50, 500, 1000.}
\label{fig: 1DLearning}
\end{figure}

In Fig. \ref{fig: 1DLearning}, we present the learning curves of the exponent $k$ and the metric $\Delta \theta$ for different memory sizes $N$. Here the minibatch size $m$ is 5 and the step size $\alpha$ is $2 \times 10^{-5}$. 
In the whole learning process, the exponent grows monotonically and the metric $\Delta \theta$ decreases monotonically to 0, indicated by Eq. (\ref{Eq: 1Dtheta}). 

We further study how the replay memory setting affects the learning performance. In practice, the total training time for an agent to learn a certain amount of knowledge is used to represent its performance. Here the total steps required to reach $k=K$ is chosen to stand for the learning ability of the agent. The value $K$ is our desired exponent value. For instance, if $K=3$, the agent is thought to learn well when its metric is $\Delta \theta = \theta - \beta_1 = 0.05 \Delta \theta^0$. Note that an agent is viewed as a better learner if it uses less training time, i.e., less learning steps, to achieve $k=K$.

First, it is always more beneficial to have a larger minibatch in this case. On the one hand, the exponent $k$ is strictly proportional to the minibatch size $m$. On the other hand, the exponent monotonically increases during the whole learning process. Thus, the larger minibatch an agent has, the faster it learns. It should be mentioned that in real experiments, the minibatch cannot be too large, cause one gradient step size is required to be small to guarantee the validity of TD-error update, and the gradient step size is proportional to the minibatch size in the defined ER algorithm.

\begin{figure}[]
\centering
\begin{subfigure}[t]{.4\textwidth}
\centering
  \includegraphics[width=0.93\linewidth]{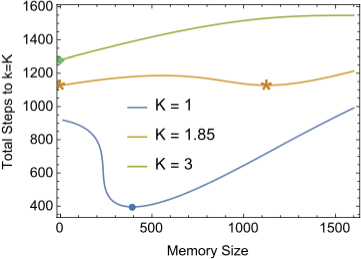}
  \caption{Steps required to target exponent $k=K$ vs. memory size.}
  \label{fig: 1DStepsVsSize}
\end{subfigure} \hfill
\begin{subfigure}[t]{.4\textwidth}
\centering
  \includegraphics[width=0.93\linewidth]{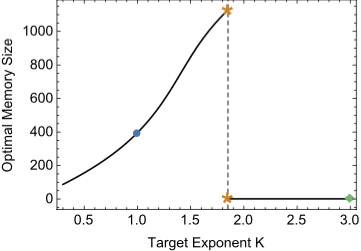}
  \caption{Optimal meomroy size vs. target exponent}
  \label{fig: 1DOptSize}
\end{subfigure}
\caption{How memory size affects performance in fixed intercept case.}
\label{fig: 1DOptN}
\end{figure}

Second, the learning ability of the agent, represented by the total steps an agent takes from the initial exponent to $k=K$, has nonmonotonic dependence on the memory size, as plotted by Fig. \ref{fig: 1DStepsVsSize}. As the learning proceeds, the weight update from the currently collected transition, i.e., $\alpha \delta_i \Delta_{\theta}Q(x_i, a_i;\theta)$, first decreases, and then grows after the moment when the state increases to be above 0.
In the same time, for any transition in this setting, the weight update is always in the correct direction towards the real value $\beta_1$. That is, a larger weight update is always preferable.

In the beginning of the game, older transitions in the memory replay contribute more to the weight update and a larger memory is more desirable. As soon as the agent reaches the region where the weight update from its current state is large enough, compared to the average of its past experience, only updating the current transition becomes a good strategy.
As illustrated by the blue curve in Fig. \ref{fig: 1DLearning}, for memory size $N=1$, it grows most slowly at start, the velocity of its exponent increases gradually surpasses other conditions with different memory size, and grows most rapidly among all possible choices of $N$ after learning step $t=1000$.
Thus, the optimal memory size increases with the growth of the target exponent $K$, abruptly falls to 1 at a certain $K$,and remains to be 1 thereafter, as demonstrated in Fig. \ref{fig: 1DOptSize}.

\begin{figure}[h]
\centering
\includegraphics[width=0.4\textwidth]{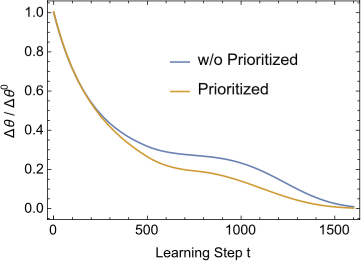}
\caption{\label{fig: 1DPriLearning} The learning curve for the metric $\Delta \theta$ under conditions with prioritized method (orange curve) and without prioritized method (blue curve) in fixed intercept case. Here the memory size is 500, batch size is 5 for illustration purpose.}
\end{figure}

The prioritized method always outperforms the uniformly selection approach in the fixed-intercept case. It is due to the fact that, a larger absolute TD-error corresponds to a larger derivative of the action-value function with respect to the weight, and furthermore it corresponds to a larger weight update. As mentioned above, a larger weight update is always beneficial. In the prioritized setting, transitions with larger TD-error value are more frequently selected, leading to a faster convergence to real weight value. It can also be demonstrated analytically as the difference between the exponent for the prioritized setting $k_{\mathrm{pri}}(t)$ and for the original one $k(t)$ is written as
\begin{equation}
\begin{aligned}
k_{\mathrm{pri}}(t) - k(t) = c(t) v^3 t^3 \bigg[ \left(2 v t + \frac{15}{4} x_0 \right)^2 & + \frac{15}{16} x_0^2 \bigg] \\ 
& (t \leqslant N)
\end{aligned}
\end{equation}
\begin{equation}
\begin{aligned}
k_{\mathrm{pri}}(t) - k(t) = c'(t) & v^3 N^3 \bigg[ \left(2 v t - \frac{15}{4} (v t + x_0) \right)^2 \\
& + \frac{15}{16} (v t + x_0)^2 \bigg] \ \ \ \ \ (t > N),
\end{aligned}
\end{equation}
where $c(t)>0$ and $c'(t)>0$. It can be easily observed that $k_{\mathrm{pri}}(t) - k(t) > 0$ at any learning step $t$, thus prioritized learning converges more quickly than the original setting, as illustrated by Fig. \ref{fig: 1DPriLearning}.

\paragraph{Fix the slope $\theta_1$} 
Another possible one-dimensional action-value function weight is to fix the slope $\theta_1$ in Eq. (\ref{Eq: Qagent}), which means
\begin{equation}
Q_{\mathrm{agent}}(x, a; \theta) = \beta_1 \cdot (x+a) + \theta
\end{equation}
With the initial weight $\theta^0$, the initial metric is given by $\Delta \theta^0 = \theta^0 - \beta_2$, considering the real action-value function is $Q_{\mathrm{real}}(x, a; \theta) = \beta_1 \cdot (x+a) + \beta_2 $.

The metric $\Delta \theta (t) = \theta (t) - \beta_2$ is obtained analytically as
\begin{equation} \label{Eq: 1Dtheta2}
\Delta \theta(t) = \Delta \theta^0 e^{-m\alpha t}
\end{equation}

Equation (\ref{Eq: 1Dtheta2}) suggests that the metric also decreases to the target value 0 exponentially, but the exponent is proportional to the learning step $t$. In strong contrast, the exponent is third-order polynomial with respect to $t$ in the fixed-intercept case. Thus, the learning is generally slower for this fixed-slope case than the fixed-intercept case. For example, when the minibatch size is 5 and the step size is $5 \times 10^{-5}$, it takes the fixed-slope agent 30000 learning steps to achieve $\Delta \theta / \Delta \theta^0 = 0.05$, while the fixed-intercept agent only needs less than 1200 steps to learn.

The learning process and results are totally independent of the initial state, velocity of state changing, and most importantly, the memory size. The learning dynamics is fully described by the step size, the minibatch size and the initial weights. This is due to the fact that all transitions are identically useful in fixed-slope situation. At any learning step, it can be easily proved that the TD-error and the weight update are the same for all transition with all possible state values. The selection of data for update no longer matters, so different replay memory settings have same performance.

Similarly, the prioritized method has the same learning results as the original setting, cause all transitions are equal in the sense of weight update. This can be confirmed by theoretical calculation, which yields that the exponent for the prioritized setting $k_{\mathrm{pri}}(t)$ and the original one $k(t)$ satisfy
\begin{equation}
k_{\mathrm{pri}}(t) \equiv k(t).
\end{equation}

\subsection{Solution for the full model}

From Eq. (\ref{Eq: dtheta1}) and Eq. (\ref{Eq: dtheta2}), the dynamic equation for the metric $\Delta \theta_1 (t)$ is given by
\begin{subequations}
\begin{equation}
\begin{aligned}
0 = & \frac{\mathrm{d^2}\Delta \theta_1(t)}{\mathrm{d}t^2} + \frac{c_{10} + c_{11} t + c_{12} t^2 + c_{13} t^3}{d_0 + d_1 t} \frac{\mathrm{d}\Delta \theta_1(t)}{\mathrm{d}t} \\ 
& + \frac{c_{00} + c_{01} t + c_{02} t^2 + c_{03} t^3}{d_0 + d_1 t} \Delta \theta_1(t) \ \ \ \ \ \ \  (t \leqslant N)
\end{aligned}
\end{equation}
\begin{equation}
\Delta \theta_1(t=0) = \theta_1^0 - \beta_1
\end{equation}
\begin{equation}
\frac{\mathrm{d}\Delta \theta_1(t=0)}{\mathrm{d}t} = -m \alpha x_0^2  ( \theta_1^0 - \beta_1) -m \alpha x_0 (\theta_2^0 - \beta_2)
\end{equation}
\end{subequations}

\begin{subequations}
\begin{equation}
\begin{aligned}
0 = & \frac{\mathrm{d^2}\Delta \theta_1(t)}{\mathrm{d}t^2} + \frac{g_{10} + g_{11} t + g_{12} t^2 + g_{13} t^3}{h_0 + h_1 t} \frac{\mathrm{d}\Delta \theta_1(t)}{\mathrm{d}t} \\
& + \frac{g_{00} + g_{01} t + g_{02} t^2}{h_0 + h_1 t} \Delta \theta_1(t) \ \ \ \ \  (t \geqslant N)
\end{aligned}
\end{equation}
\begin{equation}
\Delta \theta_1(t=N^+) = \Delta \theta_1(t=N^-)
\end{equation}
\begin{equation}
\frac{\mathrm{d}\Delta \theta_1(t=N^+)}{\mathrm{d}t} = \frac{\mathrm{d}\Delta \theta_1(t=N^-)}{\mathrm{d}t}
\end{equation}
\end{subequations}
After $\Delta \theta_1 (t)$ is obtained, another metric $\Delta \theta_2 (t)$ can be derived based on Eq. (\ref{Eq: dtheta1}). Here the parameters are

\begin{subequations}
\begin{equation}
c_{00} = 12 \alpha  m v x_0^2
\end{equation}
\begin{equation}
c_{01} = 6 \alpha  m v^2 x_0
\end{equation}
\begin{equation}
c_{02} = 2 \alpha ^2 m^2 v^2 x_0+4 \alpha  m v^3
\end{equation}
\begin{equation}
c_{03} = \alpha ^2 m^2 v^3
\end{equation}
\begin{equation}
c_{10} = 24 \alpha  m x_0^3+24 \alpha  m x_0-12 v
\end{equation}
\begin{equation}
c_{11} = 12 \alpha  m v+36 \alpha  m v x_0^2
\end{equation}
\begin{equation}
c_{12} = 20 \alpha m v^2 x_0
\end{equation}
\begin{equation}
c_{13} = 4 \alpha  m v^3
\end{equation}
\begin{equation}
d_0 =  24 x_0
\end{equation}
\begin{equation}
d_1 = 24 v
\end{equation}
\end{subequations}

\begin{subequations}
\begin{equation}
\begin{aligned}
g_{00} = & \alpha  m v \bigg[N^2 v^2 (4-\alpha  m N)+2 N v x_0 (\alpha  m N-12) \\
&+24 x_0^2\bigg]
\end{aligned}
\end{equation}
\begin{equation}
g_{01} = 2 \alpha  m v^2 \left[N v (\alpha  m N-12)+24 x_0\right]
\end{equation}
\begin{equation}
g_{02} = 24 \alpha  m v^3
\end{equation}
\begin{equation}
\begin{aligned}
g_{10} = & 4 \bigg[v \left(\alpha  m N \left(N^2 v^2+3\right)+6\right)-\alpha  m x_0 \left(5 N^2 v^2+6\right) \\
& +9 \alpha m N v x_0^2-6 \alpha  m x_0^3\bigg]
\end{aligned}
\end{equation}
\begin{equation}
g_{11} = -4 \alpha  m v \left(5 N^2 v^2-18 N v x_0+18 x_0^2+6\right)
\end{equation}
\begin{equation}
g_{12} = 36 \alpha  m v^2 \left(N v-2 x_0\right)
\end{equation}
\begin{equation}
g_{13} = 24 \alpha  m v^3
\end{equation}
\begin{equation}
h_0 =  24 x_0 - 12 N v
\end{equation}
\end{subequations}

\begin{figure}[h]
\centering
\begin{subfigure}{.43\textwidth}
  \centering
  \includegraphics[width=0.9\linewidth]{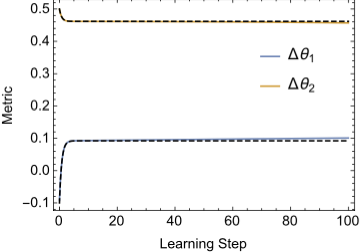}
  \caption{First stage}
  \label{fig: 2DBegining}
\end{subfigure} \hfill
\begin{subfigure}{.43\textwidth}
  \centering
  \includegraphics[width=0.9\linewidth]{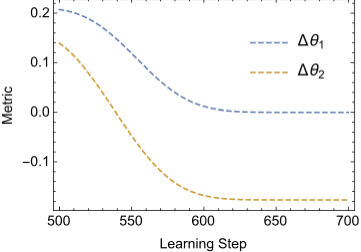}
  \caption{Last stage}
  \label{fig: 2DEnding}
\end{subfigure}
\caption{Approximated analytical solutions for two metrics. The dashed curves are analytical results from (a) Eq. (\ref{Eq: Beginning}) and (b) Eq. (\ref{Eq: Last}). (a) The solid curves represent the numerical solution.}
\label{fig: 2DAnalytical}
\end{figure}

Approximated analytical solutions can be derived. For Fig. \ref{fig: 2DLearning}, the approximated learning functions in the first and last stage are plotted in Fig. \ref{fig: 2DAnalytical}.

\begin{figure*}[h]
\centering
\begin{subfigure}[t]{.3\textwidth}
\centering
  \includegraphics[width=1\linewidth]{2D2D.png}
  \caption{ER.}
  \label{fig: 2D2D2}
\end{subfigure} \hfill
\begin{subfigure}[t]{.3\textwidth}
\centering
  \includegraphics[width=1\linewidth]{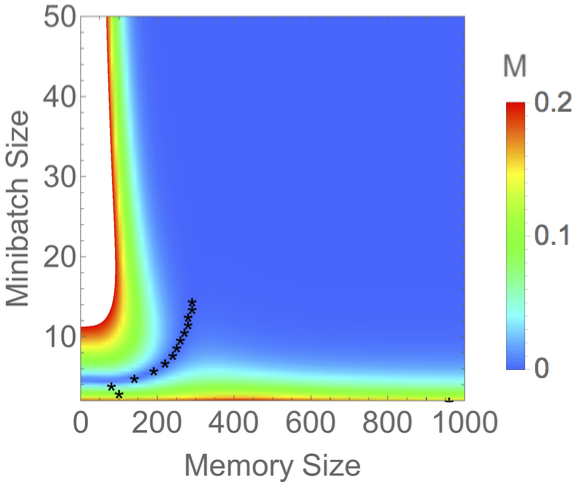}
  \caption{pER.}
  \label{fig: 2D2DPri}
\end{subfigure} \hfill
\begin{subfigure}[t]{.3\textwidth}
\centering
  \includegraphics[width=1.05\linewidth]{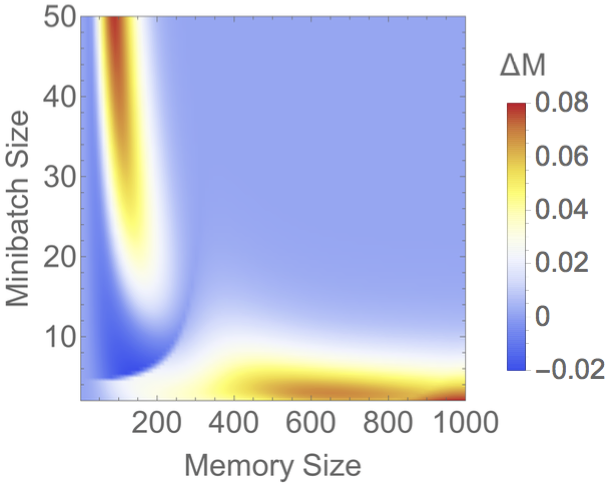}
  \caption{Compare ER and pER.}
  \label{fig: 2DComp}
\end{subfigure} \hfill
\caption{Contour plot of final absolute metric sum $M=|\Delta \theta_1 (t=1000)|+|\Delta \theta_2 (t=1000)|$ as a function of memory size and minibatch size for (a) ER and (b) pER algorithms. Smaller final total metric stands for better performance. The stars denote the optimal memory sizes given minibatch values. 
(c) The contour plot of the difference of the final absolute metric sum $M$ for the original setting and the prioritized setting as a function of memory size and minibatch size, i.e., the result of subtracting Fig. \ref{fig: 2D2D} by Fig. \ref{fig: 2D2DPri}. The positive value represents that the prioritized method is useful, while the negative value denotes that the prioritized method is harmful}
\label{fig: 2D2Ds}
\end{figure*}

In the beginning part, the metrics are estimated as
\begin{subequations} \label{Eq: Beginning}
\begin{equation}
\begin{aligned}
\Delta \theta_1 (t) = & \frac{\Delta \theta _1^0+\Delta \theta _1^0 x_0^2 e^{\alpha  m t \left(-x_0^2-1\right)}-\Delta \theta _2^0 x_0}{x_0^2+1} \\
& + \frac{\Delta \theta _2^0 x_0 e^{\alpha  m t \left(-x_0^2-1\right)}}{x_0^2+1}
\end{aligned}
\end{equation}
\begin{equation}
\begin{aligned}
\Delta \theta_2 (t) = & \frac{\Delta \theta _1^0 x_0 e^{\alpha  m t \left(-x_0^2-1\right)}+\Delta \theta _2^0 e^{\alpha  m t \left(-x_0^2-1\right)}}{x_0^2+1} \\
& + \frac{\Delta \theta _2^0 x_0^2-\Delta \theta _1^0 x_0}{x_0^2+1} 
\end{aligned}
\end{equation}
\end{subequations}
As illustrated by the dashed curve in Fig. \ref{fig: 2DBegining}, this estimation fits the real solution well when the learning step is less than 100.

Equation (\ref{Eq: Beginning}) indicates that the weights changes rapidly to
\begin{subequations} \label{Eq: BeginningConv}
\begin{equation}
\Delta \theta_1 (t) \rightarrow \Delta \theta_1^0 -\frac{x_0 \left(\Delta \theta _2^0+\Delta \theta _1^0 x_0\right)}{x_0^2+1}
\end{equation}
\begin{equation}
\Delta \theta_2 (t) \rightarrow \Delta \theta_2^0 -\frac{\Delta \theta _2^0+\Delta \theta _1^0 x_0}{x_0^2+1}
\end{equation}
\end{subequations}
After the swift change, the metrics remain constant for a short period, as shown in Fig. \ref{fig: 2DBegining}.
Note that the value changing behavior happens so fast that the values given in Eq. (\ref{Eq: BeginningConv}) do not rely on the memory size, policy, minibatch size and even the step size.

In the last stage of the learning, the metrics are approximated as
\begin{subequations} \label{Eq: Last}
\begin{equation}
\Delta \theta_1 (t) = \Delta \theta _1^1 e^{-\frac{1}{3} \alpha  m t^3 v^2}
\end{equation}
\begin{equation}
\begin{aligned}
\Delta \theta_2 (t) = & \frac{3^{2/3} \Delta \theta _1^1 \sqrt[3]{\alpha  m v^2} \Gamma \left(\frac{2}{3},\frac{1}{3} m t^3 v^2 \alpha \right)3 \Delta \theta _2^1 v}{3 v} \\
&- \frac{3^{2/3} \Delta \theta _1^1 \Gamma \left(\frac{2}{3}\right) \sqrt[3]{\alpha  m v^2}}{3 v},
\end{aligned}
\end{equation}
\end{subequations}
where $\Delta \theta _1^1$ and $\Delta \theta _2^1$ are estimated values for $\Delta \theta _1$ and $\Delta \theta _2$ when the agent enters the last stage. The learning curve in the last stage is illustrated in Fig. \ref{fig: 2DEnding}. Here the guess for $\Delta \theta _1^1$ and $\Delta \theta _2^1$ can vary a lot. In the rigorous result, we observe a delayed effect, which is due to the fact that the agent does not fully enter the last stage in our experiment time scale and the neglected terms for the derivation of Eq. (\ref{Eq: Last}) also contribute to the result.

From Eq. (\ref{Eq: Last}) we observe that the weights finally approach
\begin{subequations} \label{Eq: LastConv}
\begin{equation}
\Delta \theta_1 (t) \rightarrow 0
\end{equation}
\begin{equation} \label{Eq: LastConv2}
\Delta \theta_2 (t) \rightarrow \frac{3 \Delta \theta _2^1 v-3^{2/3} \Delta \theta _1^1 \Gamma \left(\frac{2}{3}\right) \sqrt[3]{\alpha  m v^2}}{3 v}
\end{equation}
\end{subequations}

It should be addressed that the metric $\Delta \theta_2$ does not converge to 0 because the agent changes the state unidirectionally and the movement finally reaches a balance with the weight update. Thus, in order to make the metric as near 0 as possible at last, it is crucial for the agent to have small value of Eq. (\ref{Eq: LastConv2}) when it enters the final stage of the learning process.

\begin{figure*}[h]
\centering
\begin{subfigure}[t]{.39\textwidth}
\centering
  \includegraphics[width=1\linewidth]{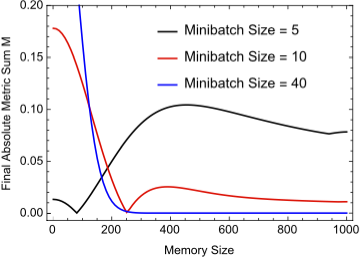}
  \caption{Final absolute metric sum $M$ vs. memory size for different minibatch size.}
  \label{fig: 2DMetricVsSize2}
\end{subfigure} \hspace*{10mm} 
\begin{subfigure}[t]{.39\textwidth}
\centering
  \includegraphics[width=1\linewidth]{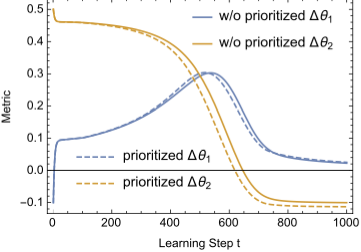}
  \caption{Memory size $N=100$, Minibatch size $m=10$.}
  \label{fig: 2DN100}
\end{subfigure}
\begin{subfigure}[t]{.39\textwidth}
\centering
  \includegraphics[width=1\linewidth]{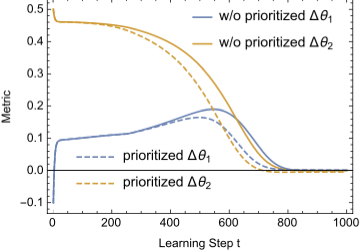}
  \caption{Memory size $N=250$, Minibatch size $m=10$.}
  \label{fig: 2DN250}
\end{subfigure} \hspace*{10mm} 
\begin{subfigure}[t]{.39\textwidth}
\centering
  \includegraphics[width=1\linewidth]{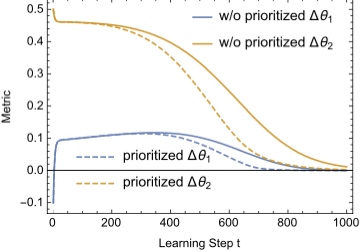}
  \caption{Memory size $N=1000$, Minibatch size $m=10$.}
  \label{fig: 2DN1000}
\end{subfigure}
\caption{(b-d) Learning curve for the two metrics $\Delta \theta_1$ and $\Delta \theta_2$ with or without prioritized methods.}
\label{fig: 2DExplain}
\end{figure*}

Here we illustrate in details how the memory setting affects the learning performance from the trade-off between the overshooting and the weight update.
When the memory size is small, overshooting is more likely to take place owing to the limited memory size. 
As depicted by the solid curves in Fig. \ref{fig: 2DN100} when the memory size is 100, the metric $\Delta \theta_2$ is gradually fitted from positive to negative, while $\Delta \theta_1$ remains to be positive after the first stage.
From Eq. (\ref{Eq: LastConv2}) we know that, in this case, the absolute value which $\Delta \theta_2$ approaches is large, confirmed by Fig. \ref{fig: 2DN100}.
With the growth of memory size, the overshooting effect is mitigated. 
In this case, the weight update is first decelerated and then slightly accelerates.
The optimal memory size is reached around 250, as depicted by Fig. \ref{fig: 2DN250}. 
When the memory size continues to increase, the agent suffers little overshooting issue. The convergence is slower than the optimal one because of the smaller weight update, as shown in Fig. \ref{fig: 2DN1000}.

\begin{figure}[]
\centering
\includegraphics[width=0.43\textwidth]{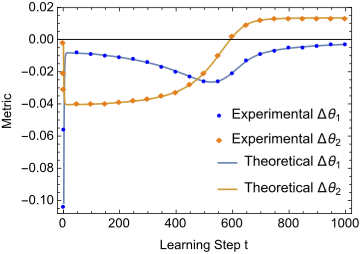}
\caption{\label{fig: 2DDiscountLearning} Learning curve for two metrics $\Delta \theta_1 = \theta_1 - \theta_1^r$ and $\Delta \theta_2 = \theta_2 - \theta_2^r$, where $\theta_{1(2)}$ and $\theta_{1(2)}^r$ are the weights of the agent and the real weights, respectively. The scattered blue dots and orange squares represent the experimental results for $\Delta \theta_1$ and $\Delta \theta_2$, based on the ER algorithm. The blue and orange curve are the theoretical solutions for $\Delta \theta_1$ and $\Delta \theta_2$.}
\end{figure}

\section{Effects of memory size in prioritized replay}

Here we analyze the effects of the memory replay on the prioritized methods. Fig. \ref{fig: 2D2DPri} presents the dependence of the final absolute metric sum $M$ on memory size and minibatch size, which exhibits a similar behavior as in the original setting and the causes are also similar. $M$ has a nonmonotonic dependence on memory given the batch size less than 15, and decreases monotonically with the increase of memory.
Fig. \ref{fig: 2DComp} depicts how the difference of $M$ between the original and prioritized settings depend on memory size and minibatch size. A positive difference value means that the prioritized setting is better. Fig. \ref{fig: 2DCompSign} is derived from it.

In principle, the pER algorithm is found to perform relatively worse than the ER when both the memory size and minibatch size is small, as indicated by Fig. \ref{fig: 2DComp}. This could also be explained with  the trade-off between the overshooting and quick weight update, similar to the situation in Fig. \ref{fig: 2DMetricVsSize2}. As plotted in Fig. \ref{fig: 2DN100} and Fig. \ref{fig: 2DN1000}, for the memory size $N$ of 100, the prioritized setting makes the overshooting even worse; while for $N=1000$, the weights are quickly updated and the prioritized agents converge faster. It should be noted that there are more complicated situations in this two-dimensional-weight situation. For $\Delta \theta_2$, prioritized scheme always results in a faster weight update, while for $\Delta \theta_1$, this does not necessarily hold true according to the definition of the weight update. For example, in Fig. \ref{fig: 2DN250}, before the learning step 500, the non-prioritized case actually learns faster than the prioritized case.

\section{Nonzero discount factor}
The theoretical model when $\gamma > 0$ also fits well with the experiments' result, as illustrated in Fig. \ref{fig: 2DDiscountLearning}, where the minibatch size $m$ is $5$, the step size $\alpha$ is $0.01$, the discount factor $\gamma$ is 0.5, the real weights $\beta_1$ and $\beta_2$ are 0.1 and 0.5, and the initial weights $\theta_1^0$ and $\theta_2^0$ are 0 and 1.

\newpage

\nocite{langley00}

\bibliography{example_paper}
\bibliographystyle{icml2017}

\end{document}